\documentclass{article} 
\usepackage{2027_conference,times}

\usepackage{amsmath,amsfonts,bm}

\def\eqref#1{equation~\ref{#1}}

\def\1{\bm{1}}

\def\vs{{\bm{s}}}

\DeclareMathAlphabet{\mathsfit}{\encodingdefault}{\sfdefault}{m}{sl}
\SetMathAlphabet{\mathsfit}{bold}{\encodingdefault}{\sfdefault}{bx}{n}

\usepackage{graphicx}
\usepackage{hyperref}
\usepackage{url}
\usepackage{booktabs}       
\usepackage{amsmath}
\usepackage{algorithm}
\usepackage{algpseudocode}
\title{UltraDub: Towards Authentic Dubbing by Unifying Visually-Steered Flow Learning and Trajectory Guidance}

\author{%
  \begin{minipage}{\textwidth}
    \centering\bfseries\small
    Gaoxiang~Cong$^{1,2}$ \quad Liang~Li$^{1}$ \quad Jianwei~Wen$^{1,2}$ \quad Zhedong~Zhang$^{3,4}$ \\
    \quad Zheng-Jun Zha$^{5}$ \quad Qingming~Huang$^{2}$\\[2pt]
    \normalfont\footnotesize
    $^{1}$Institute of Computing Technology, Chinese Academy of Sciences \quad $^{2}$University of Chinese Academy of Sciences \quad $^{3}$Hangzhou Dianzi University \quad $^{4}$ ByteDance \quad $^{5}$University of Science and Technology of China\\[2pt]
  \end{minipage}
}

\finalcopy

\begin{document}

\maketitle

\fancyhead{}
\renewcommand{\headrulewidth}{0pt}

\def\eg{\emph{e.g.}} 
\def\Eg{\emph{E.g.}}
\def\vs{\emph{v.s.}} 
\def\ie{\emph{i.e.}} 
\def\Ie{\emph{I.e.}}
\def\etc{\emph{etc.}} 
\def\wrt{\emph{w.r.t.}} 
\def\etal{\emph{et al.}}

\begin{abstract}
Visual voice cloning requires intelligible, speaker-consistent speech synchronized with visible articulation. 
However, sequential multimodal conditioning can disrupt previously established temporal and speaker cues, while imbalanced inference guidance can improve linguistic accuracy at the expense of lip synchronization. 
In this paper, we propose \textbf{UltraDub}, a \textbf{U}nifying Visually-Steered Flow \textbf{l}earning and \textbf{tra}jectory Guidance \textbf{Dub}bing framework that leverages vision in two ways: as continuous motion for multimodal context aggregation, and as structural rhythm for trajectory rectification. 
Specifically, we introduce the Motion-guided Dual-context Retrieving (MDR) module, which continually recalibrates linguistic and speaker-style retrieval through shared lip-motion query residuals, utilizing independent time-conditioned gates to regulate their contributions. 
Furthermore, we propose Rhythm-anchored Trajectory Guidance (RTG), a training-free mechanism that evaluates hierarchical multimodal corrections at a visual-only predictive midpoint, safely strengthening semantic conditioning while better preserving temporal alignment.  
Finally, we construct DiverseDub, a multi-scenario benchmark to evaluate video dubbing in the wild. 
Extensive experiments demonstrate that UltraDub achieves state-of-the-art performance across four datasets. 
Our project page is available at:
\url{https://galaxycong.github.io/ultra} 
\end{abstract}

\section{Introduction}

\begin{figure}[tb]
  \centering
  \includegraphics[width=1.0\linewidth]{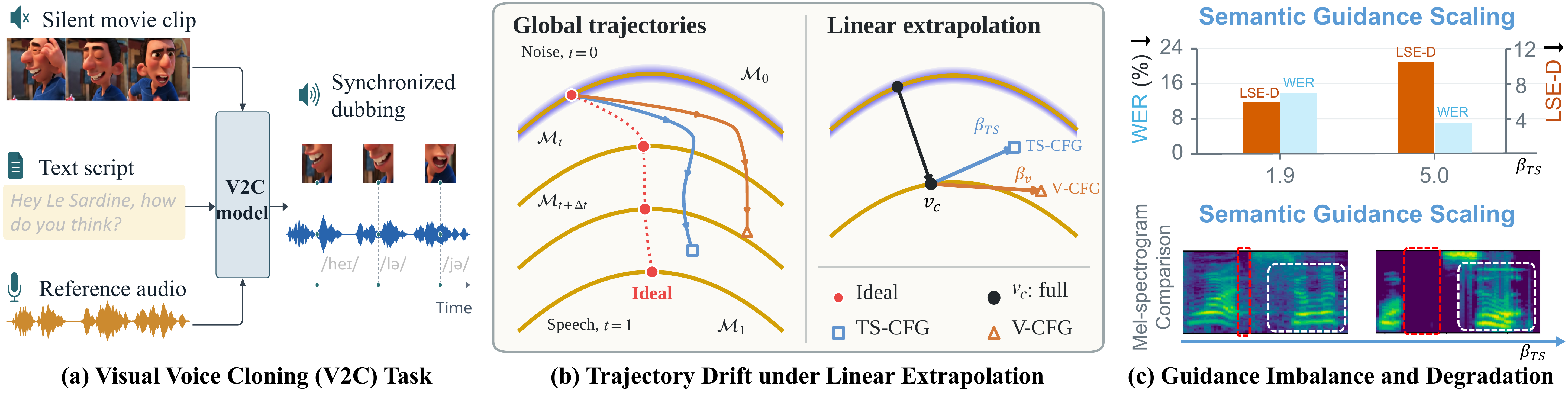}
  \vspace{-6mm} 
  \caption{
(a) Illustration of the V2C task. (b) Conventional CFG relies on linear extrapolation from fully conditioned velocity ($v_c$) via scaling factors $\beta_{*}$, making it prone to off-manifold trajectory drift. (c) For the multi-conditional V2C task, existing guidance easily triggers imbalance and degradation. 
  }
  \label{fig:intro}
  \vspace{-6mm} 
\end{figure}

Visual Voice Cloning (V2C) aims to generate high-fidelity speech
for characters in silent videos, jointly conditioned on a text
script and reference speech~\cite{chen2022v2c}.
As shown in Fig.~\ref{fig:intro}(a), each condition imposes distinct demands: the script specifies linguistic content, the reference speech conveys speaker characteristics, and the silent video defines articulation rhythm and pauses.
Driven by these capabilities, V2C demonstrates immense application potential in film post-production and personalized AIGC. 
However, achieving authentic dubbing is fundamentally more challenging than traditional text-to-speech (TTS) tasks, as it requires models to synthesize intelligible, expressive, and style-consistent speech without sacrificing fine-grained audiovisual alignment.

Existing V2C methods primarily pursue two complementary directions: expressive prosody modeling and acoustic pretraining. 
The former translates hierarchical visual cues into expressive speech (\eg, M2CI-Dubber~\cite{zhao2025towards} utilizes facial expressions and HPMDubbing~\cite{cong2023learning} further leverages scene information to extract emotion). 
However, these methods learn directly from movie corpora and may suffer from poor data quality and background interference~\cite{zhang2024speaker}, worsening clarity and Word Error Rate (WER). 
The latter direction transfers pronunciation knowledge from clean TTS corpora. For example, ProDubber~\cite{zhang2025produbber} and InstructDubber~\cite{zhang2025instructdubber} use pretrained TTS models~\cite{YinghaoStyleTTS2} to improve clarity. 
However, they rely on duration predictors coupled with these TTS architectures, which enforce rigid, integer-quantized phoneme-to-frame assignments, ultimately degrading lip synchronization and naturalness~\cite{jiang2025megatts}. 
Crucially, both paradigms rely heavily on Montreal Forced Aligner (MFA)~\cite{mcauliffe2017montreal}, which is an external forced aligner used to predefine temporal boundaries, inherently increasing pipeline complexity and error propagation.

Recent V2C methods explore joint modeling of text, video, and reference speech without explicit MFA supervision, through autoregressive codec prediction~\cite{sung2025voicecraft} or non-autoregressive flow matching. 
By jointly refining acoustic frames rather than decoding them autoregressively, flow-matching models support parallel sequence generation and improve efficiency. 
For example, AlignDiT~\cite{AlignDiT} combines conditional flow matching with auxiliary Connectionist Temporal Classification (CTC) supervision to learn implicit multimodal alignment. 
Recently, CoSyncDiT~\cite{cong2026cosyncdit} introduces a progressive architecture to sequentially perform speaker adaptation, visual calibration, and text alignment. 
However, sequential conditioning forces each stage to rely heavily on the current evolving representation, without ensuring that prior cues are preserved. 
As a result, later semantic updates may gradually deviate from the visual calibration and disrupt established speaker cues, exacerbating the risk of cross-modal interference.

Beyond architectural design, inference poses another challenge for flow-based dubbing methods, which typically rely on classifier-free guidance (CFG). 
As illustrated in Fig.~\ref{fig:intro}(b), CFG strengthens conditioning through linear extrapolation~\cite{wang2026cfg}, adding scaled residuals to the fully conditioned velocity. 
However, simply amplifying one condition may compromise adherence to others, risking off-manifold drift and artifacts. 
For example, Fig.~\ref{fig:intro}(c) demonstrates that imbalanced guidance improves clarity but sacrifices synchronization, yielding a sharp surge in LSE-D ($\downarrow$)~\cite{ChungOutofTime}. 
The accompanying mel-spectrograms corroborate this, exhibiting temporal misalignment (red boxes) and spectral distortions (white boxes). 
These observations reveal a key challenge of authentic dubbing: gains in linguistic accuracy are rendered obsolete if the speech falls out of sync with visual motion. 
Thus, effective guidance must strengthen conditional adherence while preserving audiovisual temporal alignment.

To achieve authentic dubbing, we propose \textbf{UltraDub}, a \textbf{U}nifying Visually-Steered Flow \textbf{l}earning and \textbf{tra}jectory Guidance \textbf{Dub}bing framework that naturally unifies the dual roles of vision to ensure high-fidelity dubbing, deploying it as a continuous \emph{motion} to aggregate multimodal context for velocity field estimation, and as a structural \emph{rhythm} to rectify multimodal guidance along the generative trajectory. 
Specifically, to govern multimodal conditioning during the flow learning phase, we propose the Motion-guided Dual-context Retrieving (MDR) module. 
As the hidden states evolve, MDR continually recalibrates linguistic and speaker-style retrieval through shared lip-motion query residuals and adaptively modulates their respective contributions via independent time-conditioned gates. 
With the velocity field established by MDR, we design {Rhythm-anchored Trajectory Guidance} (RTG), a training-free mechanism devised to mitigate generative drift during inference. 
It determines a visual-only predictive midpoint as a structural rhythm anchor, where hierarchical multimodal corrections are exclusively evaluated and safely injected into the fully conditioned velocity. 
Furthermore, we construct DiverseDub, a challenging V2C dubbing benchmark of 2,595 clips across eight diverse categories. Unlike existing limited datasets, it facilitates robust evaluation by incorporating unconstrained real-world scenarios with varied recording environments and speaking styles. 

Our main contributions are summarized as follows:
\begin{itemize}
    \item We propose \textbf{UltraDub}, a visually steered flow framework that unifies multimodal context retrieval and trajectory guidance for V2C. Its MDR module anchors linguistic and speaker-style retrieval to shared lip motion through independently gated context readouts.

    \item We introduce {Rhythm-anchored Trajectory Guidance (RTG)}, a training-free sampling strategy that evaluates hierarchical multimodal corrections at a visual-only predictive midpoint, incorporating visual rhythm into conditional guidance.

    \item We construct {DiverseDub}, a challenging benchmark comprising 2,595 clips across eight categories, extending V2C evaluation to unconstrained recording environments and diverse speaking styles.

    \item Experiments on three public datasets and DiverseDub demonstrate state-of-the-art performance compared with existing methods. 
    
\end{itemize}

\section{Related Work}
\label{headings}

\textbf{Visual Voice Cloning.} Visual Voice Cloning (V2C) generates speech for a target video given a script and reference speech, requiring linguistic accuracy, speaker similarity, and lip synchronization to be satisfied jointly~\cite{chen2022v2c}.
Earlier methods explicitly model visual prosody and duration: HPMDubbing and StyleDubber associate lip motion with phoneme timing, whereas ProDubber transfers acoustic knowledge from clean TTS corpora~\cite{cong2023learning,cong2024styledubber,zhang2025produbber}. 
More recently, InstructDubber integrates multimodal large language models to provide detailed duration descriptions. However, its audiovisual alignment remains sub-optimal~\cite{zhang2025instructdubber}. 
This is because its synchronization heavily relies on forced boundaries or predicted durations, allowing local timing errors to propagate into pronunciation and lip synchronization. 
Recent methods replace explicit alignment with joint generation: VoiceCraftDub predicts codec tokens autoregressively, while AlignDiT and CoSyncDiT learn implicit alignment with flow-based DiTs~\cite{sung2025voicecraft,AlignDiT,cong2026cosyncdit}. Nevertheless, these flow-based approaches still struggle with cross-modal interference and guidance imbalance.
To overcome these limitations, we propose UltraDub, which elegantly unifies the dual roles of vision by employing Motion-guided Dual-context Retrieving (MDR) for robust flow learning, and introducing Rhythm-anchored Trajectory Guidance (RTG) to rectify generative drift during inference, achieving authentic dubbing.

\textbf{Classifier-Free Guidance for Flows.} Classifier-free guidance (CFG) strengthens conditional generation by extrapolating the velocity field along the difference between conditional and unconditional predictions~\cite{ho2022classifier}. In text-to-image diffusion, while this improves prompt adherence, it can distort the sampled distribution and introduce artifacts. Thus, CFG++ replaces unconstrained extrapolation with manifold-aware guidance~\cite{chung2025cfg++}. Flow models inherit this sensitivity under deterministic ODE integration: CFG-Zero* suppresses unreliable early guidance, whereas Rectified-CFG++ anchors conditional corrections to the learned transport path~\cite{fan2025cfg, RectifiedCFG++}. 
However, these methods predominantly address a single semantic condition. In contrast, V2C is more complex, coupling linguistic content, speaker identity, and visual timing within a single trajectory. 
Amplifying text-speaker guidance may improve pronunciation, but it simultaneously alters the joint velocity governing phoneme onsets, durations, and pauses~\cite{AlignDiT}. 
Consequently, subsequent residuals are prone to being evaluated along an altered, drifted trajectory. 
In this work, we evaluate multimodal corrections at a visual rhythm anchor, ensuring that semantic refinements remain bound to the target temporal dynamics. 


\section{Method}
\label{sec:method}

\subsection{Overview}
\label{sec:overview}

\begin{figure}[tb]
  \centering
  \includegraphics[width=1.0\linewidth]{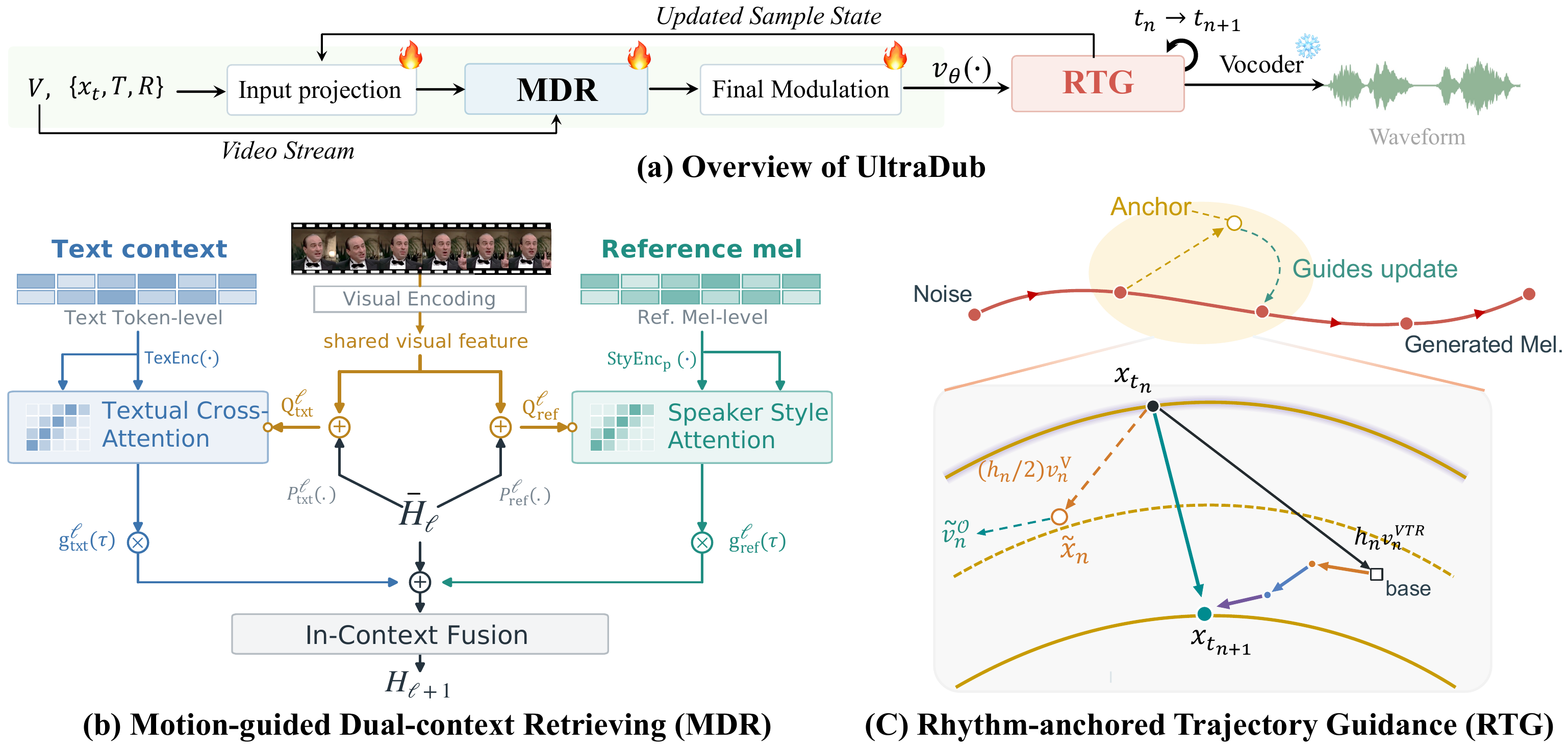}
  \vspace{-6mm} 
  \caption{Overview of UltraDub, a \textbf{U}nifying Visually Steered Flow \textbf{l}earning and \textbf{tra}jectory Guidance \textbf{Dub}bing framework. 
  The visually steered flow learning phase consists of an input projection, the core Motion-guided Dual-context Retrieving (MDR) layers, and a final modulation to predict the target velocity field $v_\theta$. 
  With the velocity field established, Rhythm-anchored Trajectory Guidance (RTG) acts as a training-free inference mechanism. It constructs a visual-only midpoint as a rhythm anchor to evaluate multimodal corrections, continuously rectifying the base transport velocity.}
  \label{fig:architecture}
  \vspace{-6mm} 
\end{figure}

Given a silent video $V$, a script $T$, and a reference audio clip $R$,
we learn a conditional velocity field
$v_\theta(x_t,t;\mathcal{C})$, with $\mathcal C=(V,T,R)$,
for noise-to-speech generation. 
During training, for a clean mel sequence $x_1$ and Gaussian noise
$x_0\sim\mathcal N(0,I)$, the intermediate state
at flow time $t\in[0,1]$ is $x_t=(1-t)x_0+t x_1$. 
Unlike previous methods, the proposed UltraDub introduces two novel designs, {Motion-guided Dual-context Retrieving} (MDR) and {Rhythm-anchored Trajectory Guidance} (RTG), as illustrated in Fig.~\ref{fig:architecture}. 
First, the input projection initializes prior representations from $x_t$, $R$, and $T$. 
Then, the proposed MDR (Fig.~\ref{fig:architecture}(b)) absorbs the input prior and dynamic lip-motion features extracted from $V$. 
Specifically, MDR enables the model to recalibrate the content and speaker-style context retrieval to a shared visual timeline while retaining independent control over each context's contribution by time-conditioned gates. 
Evolving through MDR, the output result then undergoes final modulation to predict the target velocity. 
RTG (Fig.~\ref{fig:architecture}(c)) guides multimodal corrections toward consistency with the motion rhythm in a training-free manner. 
For an integration step of size $h_n=t_{n+1}-t_n$ in flow time, RTG determines the rhythm-related velocity by constructing an auxiliary midpoint anchor.  
Multimodal corrections are evaluated at this rhythm anchor to adjust the base transport direction defined by the fully conditioned velocity. 
The resulting update is
$x_{t_{n+1}}=\Phi_{\mathrm{RTG}}^{h_n}[v_\theta]
(x_{t_n},t_n;\mathcal C)$,
where $\Phi_{\mathrm{RTG}}^{h_n}[v_\theta]$ denotes one RTG
integration step of size $h_n$ using the learned velocity
network $v_\theta$. 
Each updated state is fed back into the network until $t=1$. 
Finally, the generated mel-spectrogram is extracted and decoded into a waveform by a pretrained vocoder.

\subsection{Input Projection without Early Visual Fusion}
\label{sec:Input}

Unlike previous methods~\cite{AlignDiT, cong2026cosyncdit} that fuse visual features into the initial projection early, the input projection takes only the current noisy state $x_t$, the reference audio $R$, and the script $T$: 
\begin{equation}
    p_t
    =
    \mathcal{F}_{\mathrm{in}}
    \left(
        \left[
            x_t
            \,\|\,
            \mathbf{B} \odot \operatorname{MelEnc}(R)
            \,\|\,
            \operatorname{TexEnc}(T;\langle\mathrm{pad}\rangle)
        \right]
    \right) \in\mathbb{R}^{L\times D},
    \label{eq:input_projection}
\end{equation}
where $\operatorname{MelEnc}(\cdot)$ extracts the mel-spectrogram of $R$, $\mathbf{B}$ is a binary mask that retains reference frames~\cite{chen2025f5tts}, and $\odot$ denotes element-wise multiplication. 
The text encoder $\operatorname{TexEnc}(\cdot)$ comprises token embeddings and ConvNeXt V2 blocks~\cite{woo2023convnext}, and uses the padding symbol
$\langle\mathrm{pad}\rangle$ to extend the text sequence to
the mel-sequence length $L$. 
Here, $\|$ denotes channel-wise concatenation, and
$\mathcal{F}_{\mathrm{in}}(\cdot)$ is a learnable input projection with convolutional positional encoding.
The resulting $p_t$ serves as the initial prior without visual features.

\subsection{Motion-guided Dual-context Retrieving}
\label{sec:dual_context_dit}

\paragraph{Visual Encoding.} 
To ensure a fair comparison,
we adopt the same visual extractor pipeline $\operatorname{ViEnc}(\cdot)$~\cite{AlignDiT} with an AV-HuBERT-based~\cite{AVHUBERT} backbone to encode $V$:
\begin{equation}
  \mathbf{z}_{\mathrm{lip}}
  =
  (1-\mathbf{B})\odot
  \operatorname{Up}\bigl(\operatorname{ViEnc}(V)\bigr)
  \in\mathbb{R}^{L\times D},
  \label{eq:lip_encoder}
\end{equation}
where $\operatorname{Up}(\cdot)$ comprises two transposed-convolution
upsampling layers followed by a Conformer encoder to model
fine-grained temporal dynamics, while $1-\mathbf{B}$ retains the target region. 
As a result, $\mathbf{z}_{\mathrm{lip}}$ captures lip-motion patterns and provides visual rhythmic
cues for subsequent retrieval. 

\paragraph{Dual-Memory Retrieval.}
Rather than relying on early audiovisual fusion or staged progressive conditioning, MDR establishes a stable dual-memory retrieval mechanism that continually recalibrates linguistic content and speaker-style context to align with shared visual timing as the DiT hidden states evolve. 
Let the textual and speaker-style memories be $\mathbf{M}_{\mathrm{txt}}=\operatorname{TexEnc}(T)$ and $\mathbf{M}_{\mathrm{ref}} =\operatorname{StyEnc}_{\mathrm{p}} (\mathbf{B}\odot\operatorname{MelEnc}(R))$, where $\mathbf{M}_{\mathrm{txt}}$ is obtained from the same text encoder used for input projection but retains the text-token resolution. 
To construct $\mathbf{M}_{\mathrm{ref}}$, $\operatorname{StyEnc}_{\mathrm{p}}(\cdot)$ packs the valid reference frames according to $\mathbf{B}$ and encodes them into a dedicated speaker-style memory using two lightweight convolutional layers with LayerNorm and  SiLU activation. 
At layer $\ell$, let $\bar{\mathbf{H}}_\ell =\operatorname{AdaLN}_\ell(\mathbf{H}_\ell,\boldsymbol{\tau})$, where $\mathbf{H}_0$=$p_t$ and $\boldsymbol{\tau}$ embeds the flow time $t$. Two independently parameterized cross-attention branches retrieve linguistic content and speaker-style cues. For $c\in\{\mathrm{txt},\mathrm{ref}\}$, 
\begin{equation} 
\mathbf{r}_c^\ell = \operatorname{MHAttn}_c^\ell\!\left( \mathbf{Q}_c^\ell,\, \mathcal{K}_c^\ell(\mathbf{M}_c),\, \mathcal{V}_c^\ell(\mathbf{M}_c) \right), \quad \mathbf{Q}_c^\ell = \mathcal{P}_c^\ell(\bar{\mathbf{H}}_\ell) + \mathbf{z}_{\mathrm{lip}}, 
\label{eq:dual_memory} 
\end{equation} 
where $\mathcal{P}_c^\ell$, $\mathcal{K}_c^\ell$, and $\mathcal{V}_c^\ell$ are branch-specific learnable projections. 
Crucially, both queries are driven by the shared visual residual, enabling continuous recalibration of the two retrieval pathways as DiT hidden states evolve, and the outputs are integrated through independent time-conditioned gates $\mathbf{g}_c^\ell(\boldsymbol{\tau})\odot \mathbf{r}_c^\ell$. 
Each gate $\mathbf{g}_c^\ell(\cdot)$ is an independently parameterized, zero-initialized affine projection of $\operatorname{SiLU}(\boldsymbol{\tau})$. 
This couples content and style retrieval to a common visual timeline while preserving separate control over their contributions. 

 \paragraph{In-Context Fusion and Final Modulation.}
The in-context fusion module consists of a bidirectional self-attention and a feed-forward network, yielding contextually refined representations for the next MDR layer. 
After the final MDR layer, the final modulation module applies a linear projection with a time-adaptive layer to predict the target velocity field. 

\subsection{Rhythm-anchored Trajectory Guidance}
\label{sec:rafs}

In V2C, the conditioning modalities impose distinct demands:
text specifies linguistic content, reference speech conveys
voice characteristics, and video provides the timing.
However, stronger content or voice conditioning does not
necessarily preserve visual rhythm.
Instead of normal extrapolation, RTG addresses this tension by introducing rhythm anchoring and hierarchical multimodal corrections during training-free sampling. 

\paragraph{Rhythm Anchor.} 
Let $\mathcal O\in\{\emptyset,V,VT,VTR\}$ index the retained conditions, with $VTR$ denoting full conditioning $\mathcal C$, and define $v_n^{\mathcal O}:=v_\theta(x_{t_n},t_n;\mathcal O)$; in RTG, $v_n^V$ places the auxiliary anchor, whereas $v_n^{VTR}$ supplies the base transport direction. 
For an integration step $h_n:=t_{n+1}-t_n>0$, we evaluate
guidance at a predicted midpoint~\cite{RectifiedCFG++},
with its location determined specifically by the visual-only field:
\begin{equation}
t_n^m:=t_n+\frac{h_n}{2},
\qquad
\widetilde{x}_n\leftarrow x_{t_n}+\frac{h_n}{2}v_n^V, 
\label{eq:ud_visual_midpoint}
\end{equation}
where text and reference audio inputs are dropped for this prediction,
so explicit content and voice guidance do not determine
the anchor displacement.
The resulting half-step provides a visually directed
look-ahead state at which multimodal refinement is assessed,
making target motion the reference for subsequent corrections.

\paragraph{Hierarchical Multimodal Corrections.} 
At the anchor, we re-evaluate the same network under each condition configuration: 
\begin{equation}
\widetilde v_n^{\mathcal O}
\leftarrow v_\theta(\widetilde{x}_n,t_n^m;\mathcal O).
\label{eq:ud_midpoint_conditional}
\end{equation}
We organize these predictions into the hierarchy
$\emptyset\rightarrow V\rightarrow VT\rightarrow VTR$:
visual conditioning establishes the initial contrast,
text refines content given vision, and reference acoustics
further refine voice given both vision and text.
This ordering makes content guidance vision-conditioned
and voice guidance content-aware, rather than treating
the modalities as independent corrections:
\begin{equation}
\Delta_n^V:=\widetilde v_n^V-\widetilde v_n^\emptyset,
\quad
\Delta_n^T:=\widetilde v_n^{VT}-\widetilde v_n^V,
\quad
\Delta_n^R:=\widetilde v_n^{VTR}-\widetilde v_n^{VT}.
\label{eq:ud_conditional_contrasts}
\end{equation}
Evaluating all three contrasts at the same visual anchor
provides a common rhythm reference while allowing their
strengths to be controlled separately.

\paragraph{Guided Flow Update.} 
RTG combines the current fully conditioned velocity with the above anchored corrections to introduce rhythm guidance: 
\begin{equation}
  \widehat v_n \leftarrow v_n^{VTR}
  +\alpha(t_n)
  \left(s_V\Delta_n^V+s_T\Delta_n^T+s_R\Delta_n^R\right),
  \label{eq:ud_effective_velocity}
\end{equation}
where $s_V,s_T,s_R$ scale the conditional increments, and
$\alpha(t_n)=\lambda_{\max}t_n^\gamma$ with
$\lambda_{\max},\gamma\geq0$; for $\gamma=0$,
$\alpha(t_n)=\lambda_{\max}$ throughout.
The complete prediction--correction procedure defines
the one-step RTG integration operator:
\begin{equation}
  \Phi_{\mathrm{RTG}}^{h_n}[v_\theta]
  (x_{t_n},t_n;\mathcal C)
  :=x_{t_n}+h_n\widehat v_n.
  \label{eq:ud_RTG_update}
\end{equation}
This operator advances the acoustic state by $h_n$ in flow
time using the learned velocity network $v_\theta$.
Starting from $x_0\sim\mathcal N(0,I)$, we repeatedly apply
$x_{t_{n+1}}\leftarrow\Phi_{\mathrm{RTG}}^{h_n}[v_\theta]
(x_{t_n},t_n;\mathcal C)$ until $t=1$.
The generated target mel is then extracted and decoded
into speech by a fixed pretrained vocoder.

\section{DiverseDub Benchmark}
\label{sec:DiverseDub}

\begin{figure}[tb]
  \centering
  \includegraphics[width=1.0\linewidth]{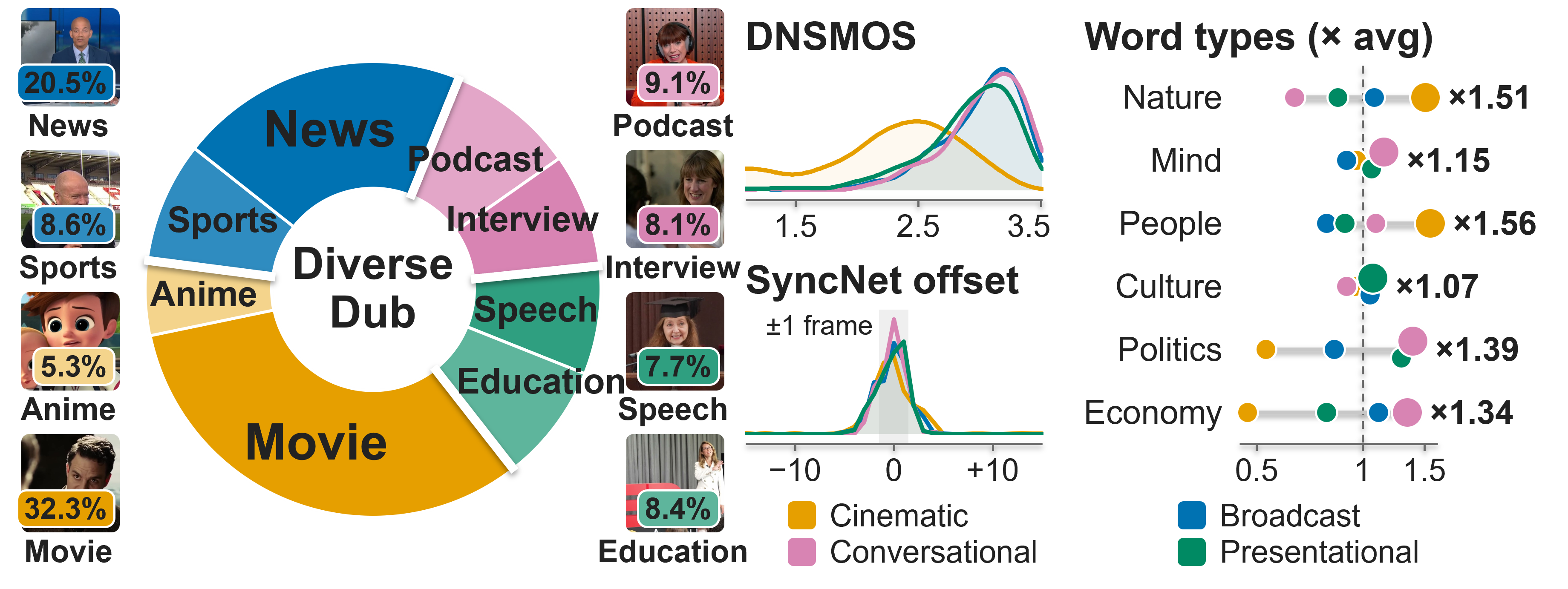}
  \vspace{-6mm} 
  \caption{Statistical overview of \textbf{DiverseDub}.
(a)~Left: distribution of clips over the eight dubbing categories, with a representative frame for each category.
(b)~Middle top: per-domain distributions of clip-level DNSMOS quality.
(c)~Middle bottom: per-domain distributions of SyncNet audiovisual offsets.
(d)~Right: relative frequency of word types among the nouns and adjectives of each domain, normalized by the corpus average.}
  \label{fig:benchmark}
  \vspace{-6mm} 
\end{figure}

To support a comprehensive evaluation under authentic dubbing conditions, we construct \textbf{DiverseDub}, an English video dubbing benchmark comprising 2,595 single-shot clips (4.79\,h) across eight dubbing categories. 
In contrast to existing benchmarks dominated by a particular source domain or production format, DiverseDub covers diverse speaking styles, visual compositions, recording environments, and editorial contexts. 
Fig.~\ref{fig:benchmark} summarizes the benchmark from four complementary perspectives: the category distribution characterizes its scenario coverage, the clip-level DNSMOS distributions describe the speech quality, the SyncNet offset distributions describe the synchronization of the retained samples, and the word-type distribution reflects its lexical diversity. 
Further details regarding DiverseDub will be provided in the Appendix.

\section{Experiments}

\textbf{Datasets.} 
We evaluate our framework on three public datasets and our newly introduced
real-world benchmark, DiverseDub.
\textbf{LRS3}~\cite{LRS3} contains TED and TEDx speech, with approximately 119,000 utterances from 5,090 speakers in its pretraining split. For evaluation, we use the LRS3-Cross test set, following the official split of AlignDiT~\cite{AlignDiT}. 
\textbf{GRID}~\cite{cooke2006audio} contains laboratory recordings with frontal faces
and a constrained vocabulary. 
The test set comprises 3,283 samples.
\textbf{CelebV-Dub}~\cite{sung2025voicecraft} is a recently released public dataset for dubbing. 
However, its evaluation capacity is highly limited in scale and diversity. 
We evaluate on 213 samples from the CelebV-Dub test set. 
\textbf{DiverseDub} extends the evaluation scope to diverse, unconstrained
real-world scenarios to assess generalization beyond existing
benchmarks (Sec.~\ref{sec:DiverseDub}). 
Across all evaluations, we adhere to the cross-utterance (\ie, Setting2) protocol: the reference audio is always sourced from a different utterance of the target speaker. 
This rigorous setting prevents information leakage and maximally tests the model's robustness for authentic dubbing.

 \begin{table}[t]
  \centering
  \caption{Results on the DiverseDub benchmark. Arrows indicate the preferred direction; the best result among dubbing
  methods is shown in bold.}
  \label{tab:diversedub_results}
  \resizebox{\linewidth}{!}{%
  \begin{tabular}{lccccc}
  \toprule
  Method & WER (\%) $\downarrow$ & SPKSIM (\%) $\uparrow$
         & EmoSIM (\%) $\uparrow$ & LSE-C $\uparrow$ & LSE-D $\downarrow$ \\
  \midrule
  GT               &  1.07 & 69.92 & 100.00 & 6.64 & 7.83 \\
  \midrule
  ProDubber        & 13.86 & 25.17 & 69.29 & 2.40 & 11.87 \\
  VoiceCraftDub   & 84.68 & 41.26 & 72.90  & 4.89 & 9.60 \\
  AlignDiT         & 20.61 & 49.78 & 80.09  & 6.33 & 8.15 \\
  InstructDubber   & \textbf{8.38} & 23.87 & 67.72  & 1.85 & 12.47 \\
  CoSyncDiT   & 12.01 & 56.32 & 81.72  & 6.88 & 7.86 \\

  \midrule
  \textbf{UltraDub (Ours)}
                   & 
                   {9.60} & \textbf{61.15} & \textbf{82.93}
                   & \textbf{7.13} & \textbf{7.56} \\
  \bottomrule
  \end{tabular}%
  }
  \end{table}

\textbf{Implementation Details.} 
UltraDub operates on mel-spectrograms extracted from 16-kHz audio and visual features from 25 FPS video streams. 
The noisy state, masked reference mel, and 512-dimensional padded text features form a 712-channel input that is projected to width 1024 without visual fusion. 
Four ConvNeXt V2 blocks produce the 512-dimensional text memory, while two kernel-3 convolution-LayerNorm-SiLU blocks encode packed reference frames into a 256-dimensional memory. 
Each MDR layer performs two parallel 16-head cross-attention reads (64 dimensions per head) from the text and reference memories. 
In both branches, the shared lip-motion residuals guide the branch-specific query projections, while keys, values, and attention normalization remain separate. 
The retrieved features are integrated through the layer's two 1024-dimensional time-conditioned gates, followed by self-attention and an FFN. 
Training masks 70--100\% of valid mel frames. 
With 32 integration steps and a sway coefficient of $-1$, the flow-time nodes are $t_n=1-\cos(\pi n/64)$, $n=0,\ldots,32$.
The rhythm anchor velocity constructs the midpoint state $\widetilde{x}_n=x_{t_n}+\frac{h_n}{2}v_n^V$, while the fully conditioned $v_n^{VTR}$ supplies the
base velocity.
At $t_n^m=(t_n+t_{n+1})/2$, four condition branches are evaluated at this rhythm anchor to obtain $\Delta_n^V$, $\Delta_n^T$, and $\Delta_n^R$.
We set $\alpha(t_n)=\lambda_{\max}t_n^\gamma$ with $(\lambda_{\max},\gamma)=(1,0)$; hence the common multiplier is one at every step, while the three corrections retain their distinct scales.

\textbf{Baselines.}
We compare our method against five recent strong baselines: ProDubber~\cite{zhang2025produbber} (CVPR'25), VoiceCraftDub~\cite{sung2025voicecraft} (ICCV'25), AlignDiT~\cite{AlignDiT} (MM'25), InstructDubber~\cite{zhang2025instructdubber} (AAAI'26), and CoSyncDiT~\cite{cong2026cosyncdit} (ECCV'26). 
To ensure a fair comparison, we use the official implementations of all baseline methods. 
Additional details are provided in Appendix~\ref{app:training_evaluation_protocol}.

  \begin{table}[t]
    \centering
    \caption{Results on the CelebV-Dub benchmark. Arrows indicate the preferred direction; the best result among dubbing
    methods is shown in bold.}
    \label{tab:CelebV-Dub_results}
    \resizebox{\linewidth}{!}{%
    \begin{tabular}{lccccc}
    \toprule
    Method & WER (\%) $\downarrow$ & SPKSIM (\%) $\uparrow$
           & EmoSIM (\%) $\uparrow$ & LSE-C $\uparrow$ & LSE-D $\downarrow$ \\
    \midrule
    GT               & 4.15 & 66.13  & 100.00 & 6.88 & 7.39 \\
    \midrule
    ProDubber        & 9.71  & 22.32 & 65.97 & 2.44 & 11.94 \\
    VoiceCraftDub   & 41.19 & 34.65 & 72.89 & 5.46 & 8.83 \\
    AlignDiT         & 25.86 & 46.25 & 74.89 & 6.29 & 8.01 \\
    InstructDubber   & \textbf{6.47} & 24.82 & 72.43 & 2.43 & 11.96 \\
    CoSyncDiT        & 12.52 & 49.05 & 78.93 & 6.86 & 7.56 \\
    \midrule
    \textbf{UltraDub (Ours)}
                     & {9.21} & \textbf{56.91} & \textbf{80.91}
                     & \textbf{7.38} & \textbf{7.09} \\
    \bottomrule
    \end{tabular}%
    }
  \end{table}

\textbf{Metrics.} We evaluate content accuracy via Word Error Rate (WER$\downarrow$)~\cite{whisper}, speaker identity via Speaker Embedding Similarity (SPKSIM$\uparrow$)~\cite{SanyuanWavLM}, emotional consistency via Emotion Embedding Similarity (EmoSIM$\uparrow$)~\cite{Ziyangemotion2vec}, and lip synchronization via SyncNet~\cite{ChungOutofTime} confidence (LSE-C$\uparrow$) and distance (LSE-D$\downarrow$).

\textbf{Results on the DiverseDub Benchmark.}
As shown in Tab.~\ref{tab:diversedub_results}, UltraDub achieves a strong balance across all evaluation metrics on the unconstrained DiverseDub benchmark.
While InstructDubber obtains the lowest WER of 8.38\%, it exhibits notable trade-offs in speaker similarity and synchronization, with an SPKSIM of 23.87\% and an LSE-D of 12.47.
In contrast, UltraDub maintains competitive intelligibility with a WER of 9.60\%, alongside stronger identity preservation and lip synchronization.
Furthermore, compared with CoSyncDiT's sequential conditioning, UltraDub improves SPKSIM from 56.32\% to 61.15\% and EmoSIM from 81.72\% to 82.93\%.
These gains are consistent with MDR's design of coordinating linguistic and speaker-style retrieval through shared visual cues.
UltraDub also achieves the best synchronization, with an LSE-C of 7.13 and an LSE-D of 7.56, supporting the use of visual timing to organize both context retrieval and RTG-based trajectory correction.

\begin{figure}[tb]
  \centering
  \vspace{-4mm} 
  \includegraphics[width=1.0\linewidth]{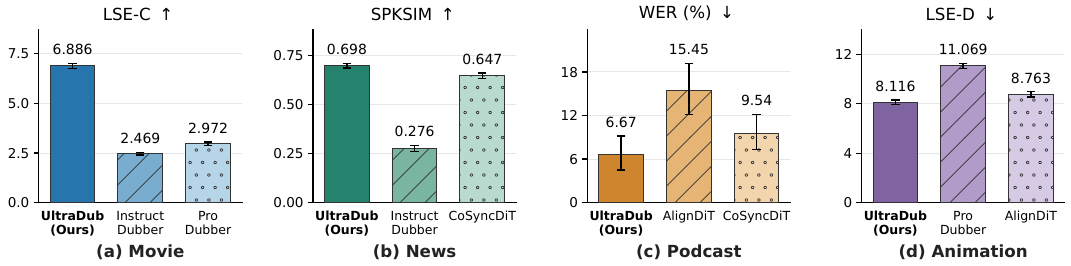}
  \caption{Category-wise comparisons on DiverseDub. Error bars indicate 95\% confidence intervals.}
  \label{fig:DiverseDub}
  \vspace{-4mm} 
\end{figure}

\noindent\textbf{Category-wise Robustness on DiverseDub.}
To further assess performance across production formats, we examine four distinct categories in Fig.~\ref{fig:DiverseDub}.
In movies and animation, UltraDub demonstrates superior synchronization, achieving an LSE-C of 6.886 and an LSE-D of 8.116, respectively, outperforming the displayed baselines, including the MFA-dependent ProDubber.
For news, UltraDub achieves an SPKSIM of 69.8\%, surpassing CoSyncDiT's 64.7\% and InstructDubber's 27.6\%.
For podcasts, it achieves the lowest WER among the displayed methods at 6.67\%, compared with 15.45\% for AlignDiT and 9.54\% for CoSyncDiT.
These category-level results demonstrate UltraDub's strengths in synchronization, speaker preservation, and intelligibility across diverse dubbing scenarios.

 \begin{table}[t]
    \centering
    \caption{Comparison with dubbing methods on the LRS3 benchmark. Best results among dubbing methods are shown in bold. }
    \label{tab:lrs3_cross_results}
    \begin{tabular}{l|cccc}
      \toprule
      Methods & WER (\%) $\downarrow$ & SPKSIM (\%) $\uparrow$
              & LSE-C $\uparrow$ & LSE-D $\downarrow$ \\
      \midrule
      GT                & 2.33  & 56.18 & 7.80 & 6.85 \\
      \midrule
      ProDubber         & 3.41  & 24.27 & 4.27 & 10.12 \\
      VoiceCraftDub     & 14.85 & 35.17 & 6.49 & 8.08 \\
    AlignDiT          & 1.40  & 51.46 & 7.91 & 6.78 \\
      InstructDubber    & 2.39  & 24.16 & 3.29 & 11.12 \\
      CoSyncDiT    & 2.09  & 47.53 & 7.83 & 6.88 \\
      \midrule
      \textbf{UltraDub (Ours)}
                        & \textbf{1.37} & \textbf{52.05}
                        & \textbf{8.23} & \textbf{6.50} \\
      \bottomrule
    \end{tabular}
  \end{table}

\textbf{Results on the CelebV-Dub Benchmark.} 
As shown in Tab.~\ref{tab:CelebV-Dub_results}, 
our method achieves the best speaker similarity, emotion similarity, and lip synchronization among the compared dubbing methods. Compared with CoSyncDiT, it increases SPKSIM from 49.05\% to 56.91\% and EmoSIM from 78.93\% to 80.91\%, while improving LSE-C from 6.86 to 7.38, reducing LSE-D from 7.56 to 7.09, and lowering WER from 12.52\% to 9.21\%. Although InstructDubber yields lower WER, its substantially weaker lip synchronization underscores the need to coordinate linguistic accuracy with visual timing. These results support UltraDub's visually steered design, which anchors context retrieval and trajectory correction to lip motion, jointly improving alignment and voice fidelity while retaining competitive intelligibility.

\textbf{Results on the LRS3 Benchmark.} 
As shown in Tab.~\ref{tab:lrs3_cross_results}, 
UltraDub achieves state-of-the-art performance across all four
evaluated metrics among the compared dubbing methods.
Compared with AlignDiT, the strongest baseline on all metrics,
UltraDub reduces WER from 1.40\% to 1.37\% and improves
SPKSIM from 51.46\% to 52.05\%.
The gains are more pronounced in synchronization:
LSE-C increases from 7.91 to 8.23, while LSE-D decreases
from 6.78 to 6.50.
These results show that UltraDub improves audiovisual alignment alongside linguistic accuracy and speaker similarity, supporting our approach of using visual motion to coordinate multimodal retrieval and guide the rhythm-anchored trajectory.

\textbf{Results on the GRID Benchmark.}
As shown in Tab.~\ref{tab:1grid_cross_results}, UltraDub
outperforms all compared dubbing methods across the four metrics.
It achieves a WER of 14.53\%, approaching the 13.14\% measured
on ground-truth speech, and the highest SPKSIM of 58.01\%.
UltraDub also achieves the best synchronization, with an LSE-C
of 7.60 and an LSE-D of 6.44, demonstrating joint improvements
in intelligibility, speaker preservation, and audiovisual alignment.

\begin{figure}[tb]
  \centering
  \vspace{-6mm} 
  
  \includegraphics[width=1.0\linewidth]{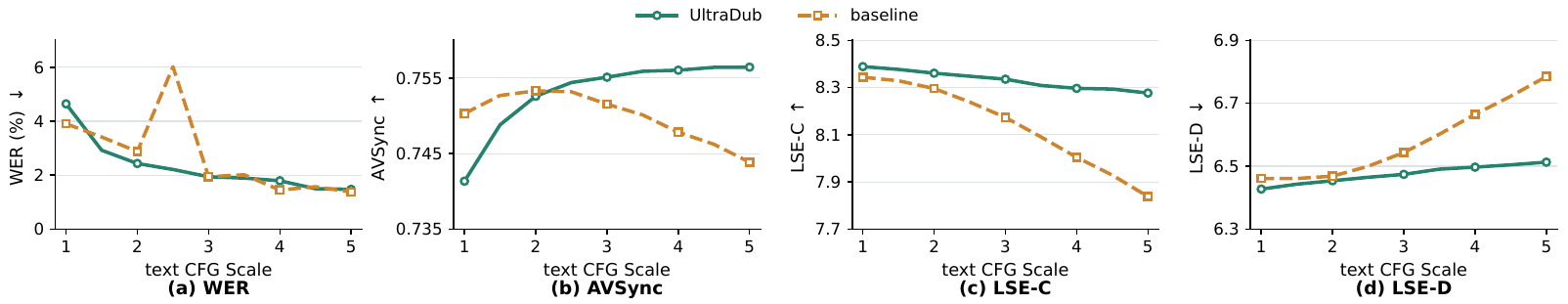}
  \caption{Preserving audiovisual alignment under increasing guidance strength on the LRS3 test set. 
  AVSync is a recently introduced metric that provides a more reliable assessment of audiovisual synchronization~\cite{AlignDiT}.}
  \vspace{-6mm} 
  \label{fig:joint}
\end{figure}

 \begin{table}[t]
        \centering
        \caption{Comparison with dubbing methods on the GRID benchmark.
        EmoSIM is not evaluated because GRID consists primarily of neutral emotion.}
        \label{tab:1grid_cross_results}
        \begin{tabular}{l|cccc}
          \toprule
          Methods & WER (\%) $\downarrow$ & SPKSIM (\%) $\uparrow$
                  & LSE-C $\uparrow$ & LSE-D $\downarrow$ \\
          \midrule
          GT                 & 13.14 & 63.69 & 7.14 & 6.84 \\
          \midrule
          ProDubber          & 44.29 & 34.91 & 3.71 & 11.08 \\
          VoiceCraftDub      & 64.47 & 33.37 & 5.13 & 8.50 \\
          AlignDiT           & 19.47 & 55.06 & 7.17 & 6.84 \\
          InstructDubber     & 27.67 & 36.30 & 3.33 & 11.32 \\
          CoSyncDiT          & 19.64 & 53.45 & 7.21 & 6.86 \\
          \midrule
          \textbf{UltraDub (Ours)}
                             & \textbf{14.53} & \textbf{58.01}
                             & \textbf{7.60} & \textbf{6.44} \\
          \bottomrule
        \end{tabular}
        \vspace{-3mm} 
    \end{table}

\paragraph{Robustness to guidance strength.} 
Fig.~\ref{fig:joint} examines linguistic accuracy and audiovisual
synchronization as the scale increases. 
We use the conventional guidance strategy adopted in the official AlignDiT implementation~\cite{AlignDiT} as the baseline. 
Besides, we report AVSync~\cite{AlignDiT}, which measures the cosine similarity between AV-HuBERT features of the same
video paired with synthesized and ground-truth speech;
higher scores indicate better synchronization.
Increasing the scale from 1 to 5 reduces UltraDub's WER
from 4.65\% to 1.48\% while improving AVSync from 0.7414
to 0.7565.
By contrast, the baseline exhibits greater WER fluctuations,
and its AVSync declines after peaking near scale 2.
Although both methods show decreasing LSE-C and increasing
LSE-D, UltraDub degrades more gradually and consistently
outperforms the baseline on both metrics.
At scale 5, UltraDub yields only slightly higher WER,
1.48\% versus 1.38\%, while achieving better AVSync,
0.7565 versus 0.7439, higher LSE-C, 8.277 versus 7.838,
and lower LSE-D, 6.513 versus 6.785.
These results support the rationale behind rhythm-anchored
guidance: evaluating multimodal corrections at a shared
visual anchor helps accommodate stronger text conditioning
while limiting synchronization degradation.

\section{Conclusion}
In this paper, we propose UltraDub  to coordinate multimodal context retrieval through MDR and
anchor training-free trajectory guidance through RTG.
Together, these components establish visual timing as a shared
reference for flow learning and sampling.
We also introduce DiverseDub, a benchmark spanning eight
categories for evaluating dubbing in unconstrained real-world
scenarios.
Experiments on four benchmarks demonstrate improvements in
intelligibility, speaker preservation, and lip synchronization.
Further analysis shows that UltraDub limits alignment
degradation under stronger guidance, supporting more
balanced multimodal conditioning during synthesis.

\bibliography{2027_conference}
\bibliographystyle{2027_conference}

\clearpage
\appendix
\section{Appendix}
\label{app:appendix}

This appendix provides additional methodological, benchmark, and experimental details that complement the main paper. It includes:

\begin{itemize}
    \item \textbf{Training and Evaluation Protocol} 
    Appendix~\ref{app:training_evaluation_protocol} describes the training setting and zero-shot evaluation protocol.

    \item \textbf{Additional Details of RTG} 
    Appendix~\ref{app:rtg8} presents the complete sampling procedure and additional analysis of RTG.

    \item \textbf{Details of the Training Objective:}
    Appendix~\ref{app:Training} describes the training loss details.

    \item \textbf{Benchmark Details and Construction} 
    Appendix~\ref{Benchmark} describes the data sources, processing pipeline, category organization, transcription procedure, and quality-control criteria of DiverseDub.

    \item \textbf{Mel-Spectrogram Comparisons} 
    Appendix~\ref{Spectrogram} provides qualitative comparisons of temporal alignment and local acoustic details across different methods.

    \item \textbf{More Category-Wise Results on DiverseDub} 
    Appendix~\ref{Category} presents detailed evaluations across the eight real-world categories included in DiverseDub.

    \item \textbf{Ablation on Flow Guidance Strategies} 
    Appendix~\ref{AblationGuidance} compares RTG with conventional classifier-free guidance under increasing text-guidance strengths.

    \item \textbf{Ablation on UltraDub Components} 
    Appendix~\ref{AblationComponents} analyzes the individual contributions of MDR, its dual-memory design, and RTG.

    \item \textbf{Additional AVSync Results} 
    Appendix~\ref{AdditionalAVSync} provides supplementary evaluations of audiovisual consistency.

    \item \textbf{Subjective MOS Evaluation} 
    Appendix~\ref{SubjectiveMOS} reports human evaluations of speech naturalness and speaker similarity.

    \item \textbf{Limitations and Social Impact} 
    Appendix~\ref{Limitations} discusses potential applications, misuse risks, and considerations for responsible deployment.
\end{itemize}

\subsection{Training and Evaluation Protocol}
\label{app:training_evaluation_protocol}
AlignDiT is a strong existing baseline whose official training protocol uses LRS3. We directly use its best-performing official LRS3 checkpoint\footnote{\url{https://github.com/kaistmm/AlignDiT}}, without retraining it or adapting it to any target domain. 
Following this setup, we train UltraDub and other competing methods on the LRS3 to ensure a fair comparison. 
The large scale of LRS3 provides sufficient training data for a reliable comparison across methods, while CelebV-Dub, GRID, and DiverseDub are reserved for zero-shot evaluation. No samples from these three datasets are used for training or dataset-specific fine-tuning. At inference time, we evaluate all methods on the LRS3 test set and directly apply them to the CelebV-Dub and GRID test sets, as well as our proposed DiverseDub benchmark. The results on CelebV-Dub, GRID, and DiverseDub therefore measure zero-shot cross-dataset generalization beyond LRS3.

\subsection{Additional Details of Rhythm-anchored Trajectory Guidance}
\label{app:rtg8}

RTG separates two decisions in multimodal sampling: \emph{where} to evaluate conditional refinements and \emph{how strongly} to apply each refinement. 
It evaluates multimodal corrections at a visual rhythm anchor.
We present the complete sampling procedure in Algorithm~\ref{alg:rtg8}
and provide further analysis of RTG below, using the notation from the
main paper throughout. 

Algorithm~\ref{alg:rtg8} expands
$\Phi_{\mathrm{RTG}}^{h_n}[v_\theta]$ into a complete noise-to-speech
sampling procedure from $t=0$ to $t=1$. 
Here $\operatorname{EvalBranches}(x,t;\mathcal C)$
evaluates the same network in the order $(VTR,VT,V,\emptyset)$ in one
packed call. The $VT$ branch drops reference acoustics, the $V$ branch
additionally drops text, and the unconditional branch drops all three
modalities. 
Following the main paper, for
$\mathcal O\in\{\emptyset,V,VT,VTR\}$, write
$v_n^{\mathcal O}:=v_\theta(x_{t_n},t_n;\mathcal O)$ for the current-state
prediction. 
The corresponding predictions at the rhythm anchor are
\begin{equation}
 \widetilde v_n^{\mathcal O}
 \leftarrow v_\theta(\widetilde x_n,t_n^m;\mathcal O).
\label{eq:rtg8_midpoint_conditional}
\end{equation}
Line~\ref{line:rtg8_midpoint} evaluates these four predictions in one packed
call. 
Their differences in lines~\ref{line:rtg8_visual_increment}--\ref{line:rtg8_reference_increment}
follow the hierarchy $\emptyset\rightarrow V\rightarrow VT\rightarrow VTR$:
$\Delta_n^V$ adds vision, $\Delta_n^T$ adds text given vision, and
$\Delta_n^R$ adds reference acoustics given vision and text. 

\begin{algorithm}[htbp]
\caption{Rhythm-anchored Trajectory Guidance (RTG)}
\label{alg:rtg8}
\begin{algorithmic}[1]
\Require Trained $v_\theta$; $\mathcal C=(V,T,R)$; reference mask $\mathbf B$ and valid-frame masks;
\Statex \hspace{\algorithmicindent} grid $0=t_0<\cdots<t_N=1$; scales $s_V,s_T,s_R$;
\Statex \hspace{\algorithmicindent} $\lambda_{\max},\gamma\ge0$; fixed vocoder $\mathcal D$
\State Sample $x_0\sim\mathcal N(0,I)$ on valid mel frames; set padding to zero
\For{$n=0,\ldots,N-1$}
 \State $h_n\gets t_{n+1}-t_n$; $t_n^m\gets t_n+h_n/2$
 \State $(v_n^{VTR},v_n^{VT},v_n^V,v_n^\emptyset)
       \gets\operatorname{EvalBranches}(x_{t_n},t_n;\mathcal C)$ \label{line:rtg8_current}
 \State $\widetilde x_n\gets x_{t_n}+(h_n/2)v_n^V$ \Comment{Visual rhythm anchor} \label{line:rtg8_anchor}
 \State $\widetilde v_n^{\mathcal O}\gets v_\theta(\widetilde x_n,t_n^m;\mathcal O),
       \quad\forall\mathcal O\in\{\emptyset,V,VT,VTR\}$ \Comment{One packed call} \label{line:rtg8_midpoint}
 \State $\Delta_n^V\gets\widetilde v_n^V-\widetilde v_n^\emptyset$ \label{line:rtg8_visual_increment}
 \State $\Delta_n^T\gets\widetilde v_n^{VT}-\widetilde v_n^V$
 \State $\Delta_n^R\gets\widetilde v_n^{VTR}-\widetilde v_n^{VT}$ \label{line:rtg8_reference_increment}
 \If{$\gamma=0$}
  \State $\alpha(t_n)\gets\lambda_{\max}$
 \Else
  \State $\alpha(t_n)\gets\lambda_{\max}t_n^\gamma$
 \EndIf
 \State $\widehat v_n\gets v_n^{VTR}
       +\alpha(t_n)(s_V\Delta_n^V+s_T\Delta_n^T+s_R\Delta_n^R)$ \label{line:rtg8_velocity}
 \State $x_{t_{n+1}}\gets x_{t_n}+h_n\widehat v_n$ \label{line:rtg8_update}
\EndFor
\State $x_{\mathrm{out}}\gets\mathbf B\odot\operatorname{MelEnc}(R)+(1-\mathbf B)\odot x_{t_N}$
\State \Return $\mathcal D(\operatorname{TargetCrop}(x_{\mathrm{out}}))$
\end{algorithmic}
\end{algorithm}

The base velocity and $\alpha$ are evaluated at $t_n$, not $t_n^m$. 
Reference frames are restored after integration, rather than clamped at each step. 
No additional noise is injected at the midpoint. 
With the default $N=32$ and sway coefficient $-1$, the nonuniform grid is $t_n=1-\cos(\pi n/64)$, $n=0,\ldots,32$. The midpoint time is $(t_n+t_{n+1})/2$, computed \emph{after} the grid transformation. 
In general, UltraDub leverages vision in two complementary ways: as \emph{continuous motion} for multimodal context aggregation in MDR, and as \emph{structural rhythm} for trajectory rectification in RTG. Specifically, continuous motion updates the MDR hidden state through dual-context retrieval, while structural rhythm determines the visually guided evolution of the inference trajectory. 
In RTG, Line~\ref{line:rtg8_anchor} constructs the visual rhythm anchor by advancing the current state by a half step along the visual velocity $v_n^V$. 
This construction yields a look-ahead state at which all multimodal corrections are evaluated, ensuring that they respond to the visually directed prediction.  

\subsection{Details of Training Loss}
\label{app:Training}

UltraDub is trained with the same flow matching loss and auxiliary connectionist temporal classification (CTC) loss as AlignDiT~\cite{AlignDiT}, ensuring a fair comparison under a consistent training objective. 
Unlike the recent state-of-the-art CoSyncDiT~\cite{cong2026cosyncdit}, UltraDub requires no additional AV-HuBERT-based contrastive loss, resulting in a simpler training objective and avoiding potential over-optimization of AVSync.

\paragraph{Flow matching loss.}
Given a clean mel sequence $x_1$ and its conditions $\mathcal C=(V,T,R)$,
we sample $x_0\sim\mathcal N(0,I)$ and $t\sim\mathcal U(0,1)$,
and construct $x_t=(1-t)x_0+t x_1$ as in the main paper.
The corresponding target velocity is $x_1-x_0$.
We learn $v_\theta$ using the conditional flow-matching objective:
\begin{equation}
\mathcal L_{\mathrm{FM}}
=\mathbb E_{(x_1,\mathcal C),\,x_0,\,t}
\left[
\left\|v_\theta(x_t,t;\mathcal C)-(x_1-x_0)\right\|_2^2
\right],
\label{eq:training_fm}
\end{equation}
where the expectation is taken over training data, Gaussian noise, and
flow time. In implementation, modality dropout is applied to $\mathcal C$,
and the squared regression error is averaged over valid target mel
elements only, excluding reference frames retained by $\mathbf B$ and
padding. 
$\theta$ denotes the trainable parameters of the visually steered flow learning phase, which consists of an input projection without early visual fusion, a stack of Motion-guided Dual-context Retrieving (MDR) layers, and a final modulation module that predicts the target velocity field, as detailed in Sections~\ref{sec:Input} and~\ref{sec:dual_context_dit}.

\paragraph{Auxiliary CTC loss.}
Auxiliary connectionist temporal classification (CTC) heads predict the complete transcript $T$ by marginalizing over all valid token alignments. 
Let
$\mathcal L_{\mathrm{CTC}}$ denote the weighted mean of these CTC losses. 
The overall training objective is
\begin{equation}
\mathcal L_{\mathrm{train}}
=\mathcal L_{\mathrm{FM}}+\lambda_{\mathrm{CTC}}\mathcal L_{\mathrm{CTC}},
\label{eq:training_total}
\end{equation}
where $\lambda_{\mathrm{CTC}}$ balances the auxiliary supervision.
Training optimizes $v_\theta$, whereas Algorithm~\ref{alg:rtg8} constructs
$\widehat v_n$ from its conditional predictions only at inference.
The CTC heads are not used during sampling.

\subsection{Benchmark Details and Construction}
\label{Benchmark}
\paragraph{Data Sources and Cleaning.}

We collect English audiovisual content from YouTube using category-specific queries and curated movie and animation playlists designed to cover diverse production formats and real-world dubbing conditions. Each source video is first segmented into individual shots using pySceneDetect~\cite{castellano2026pyscenedetect}. We then detect and track visible faces, identify the active speaker by associating face tracks with the corresponding speech using TalkNet~\cite{tao2021someone}, and crop the resulting tracks into speaker-centered clips. Segments containing shot transitions, multiple active speakers, overlapping speech, within-clip speaker changes, or unreliable face--speech associations are removed. 

\paragraph{Dubbing Categories}
We assign each clip a single primary category according to the production format or dominant editorial context of its source video, and group the categories into four domains: \emph{Cinematic} (\emph{Movie}, \emph{Animation}), \emph{Broadcast} (\emph{News}, \emph{Sports}), \emph{Conversational} (\emph{Interview}, \emph{Podcast}), and \emph{Presentational} (\emph{Speech}, \emph{Education}). These categories cover scripted and animated productions, broadcast and event coverage, conversational programs, and public and educational speech. In particular, \emph{Movie} includes live-action films and film-related media, while \emph{Sports} primarily contains condensed sports and event coverage. \emph{Podcast} also includes moderated multi-participant programs and documentary-style discussions, although each retained clip contains only one active speaker. The resulting distribution, shown in Fig.~\ref{fig:benchmark}(a), provides broad coverage without allowing a single content domain to dominate the benchmark.

\paragraph{Construction Pipeline}
Following structural cleaning, we perform evaluation-oriented curation
to ensure both broad scenario coverage and reliable ground-truth
annotations. Candidates are selected across all eight categories while
avoiding excessive concentration in any individual source domain. We
retain only clips with reliable audiovisual synchronization and
adequate DNSMOS~\cite{reddy2021dnsmos} speech quality. We additionally
compute
SyncNet~\cite{ChungOutofTime} statistics for every retained clip and report
their complete per-domain distributions in Fig.~\ref{fig:benchmark},
rather than reporting only aggregate averages. The mean DNSMOS is 2.73, with
domain means ranging from 2.31 to 3.03, and 63.7\% of clips have offsets
within $\pm$1 frame.

\begin{figure}[tb]
  \centering
  \includegraphics[width=1.0\linewidth]{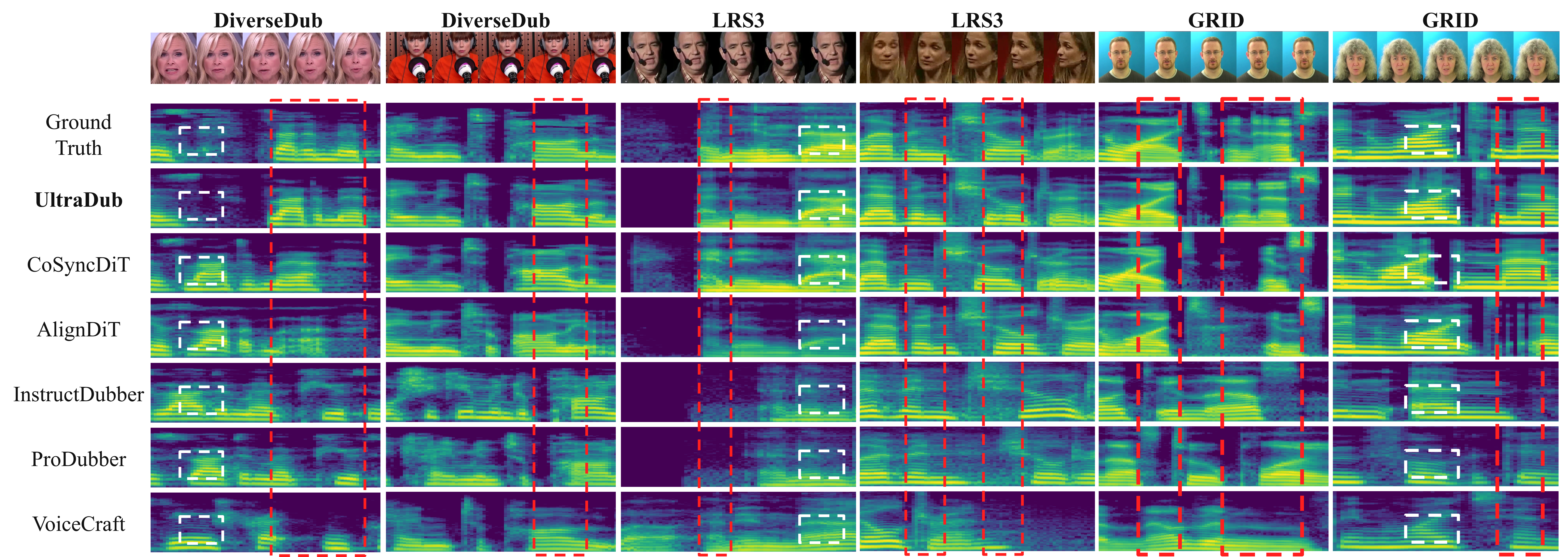}
  \caption{Mel-spectrogram comparisons between ground-truth speech and
speech synthesized by different models on DiverseDub, LRS3, and GRID.
The red and white rectangles highlight audiovisual alignment intervals
and local phonetic details, respectively.} 
  \label{fig:mel_comparision}
\end{figure}

\begin{figure}[!tb]
  \centering
  \includegraphics[width=1.0\linewidth]{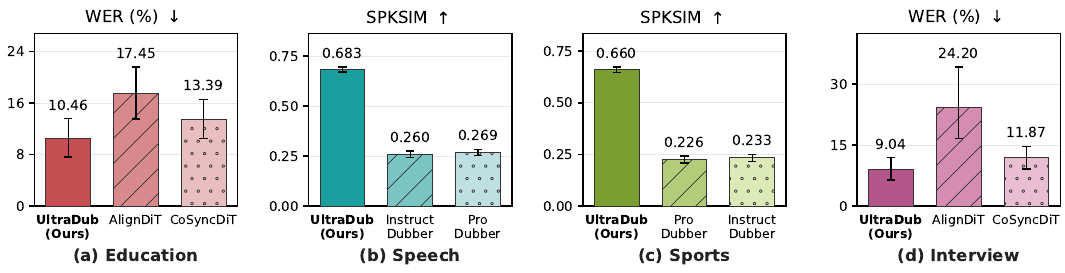}
  \caption{More category-wise comparisons on DiverseDub. Error bars indicate 95\% confidence intervals. Further details and qualitative examples are provided in the supplementary demo.}
  \label{fig:DiverseDub2}
\end{figure}

For transcript construction, Whisper large-v3~\cite{whisper} first
produces an initial transcription. Segments with low ASR confidence,
measured by the negative log-likelihood of the transcription given
the speech, are discarded, and incomplete sentences at segment boundaries
are trimmed. Each resulting clip is re-transcribed and retained only if
its WER falls below a strict threshold. An LLM agent then cross-checks the
transcriptions and corrects evident recognition errors, followed by manual
verification. Samples with unresolved transcript ambiguities, inconsistent
speaker identity, synchronization failures, or uncertain textual
annotations are discarded. The word-type distribution in Fig.~\ref{fig:benchmark}(d) is
computed from the final verified transcripts using USAS~\cite{rayson2004usas}
semantic fields and provides a complementary view of the linguistic
content covered by DiverseDub.

\subsection{Mel-Spectrogram Comparisons}
\label{Spectrogram} 
Fig.~\ref{fig:mel_comparision} presents qualitative mel-spectrogram comparisons, with red and white boxes highlighting alignment intervals (onsets, offsets, pauses) and local phonetic details, respectively. Across all benchmarks, UltraDub produces temporal boundaries and alignment trajectories that most closely resemble the ground truth, precisely synchronizing vocal events with lip movements. Although AlignDiT and CoSyncDiT generally capture the coarse timing of speech, they both exhibit local boundary shifts, delayed onsets, and inaccurate syllable offsets within the highlighted regions. 
The discrepancies are substantially more pronounced for InstructDubber,
ProDubber, and VoiceCraftDub, which trigger speech events at completely incorrect time steps and lose temporal alignment altogether.

\subsection{More Category-wise Results on DiverseDub} 
\label{Category}
Figs.~\ref{fig:DiverseDub} and~\ref{fig:DiverseDub2} report detailed results across eight real-world categories. The category-wise comparisons illustrate UltraDub's advantages in synchronization, speaker similarity, and content accuracy across diverse scenarios. 
In particular, it achieves substantially higher LSE-C on Movie and higher SPKSIM on News, Speech, and Sports, demonstrating improved lip synchronization and speaker-identity preservation across diverse content. 
UltraDub also obtains the lowest WER on Podcast (6.67\%), Education (10.46\%), and Interview (9.04\%), indicating more reliable linguistic generation. 
On Animation, it reduces LSE-D to 8.116, outperforming both AlignDiT and ProDubber. 
These consistent category-level improvements demonstrate that UltraDub generalizes robustly across heterogeneous speaking styles and visual scenarios.

\begin{figure}[tb]
  \centering
  \includegraphics[width=1.0\linewidth]{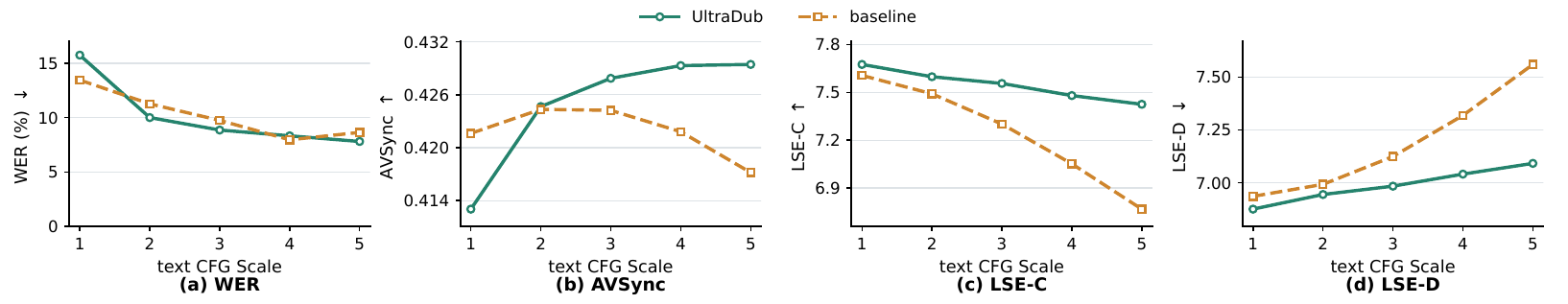}
  \caption{Preserving audiovisual alignment under increasing guidance strength on the CelebV-Dub test set. 
  AVSync is a recently introduced metric that provides a more reliable assessment of audiovisual synchronization~\cite{AlignDiT}.}
  \label{fig:joint_celeb}
\end{figure} 

 \begin{table}[!t]
    \centering
    \caption{Component ablation study of UltraDub on LRS3-Cross. 
Starting from a self-attention and feed-forward network (SA+FFN) backbone, we evaluate different MDR memory configurations and the contribution of RTG. 
Arrows indicate the preferred metric directions, and bold denotes the best  result. 
Each architectural variant is trained independently under the same training configuration, while RTG is applied only at inference time.}
    \label{tab:lrs3_ab}
    \begin{tabular}{l|cccc}
      \toprule
      Methods & WER (\%) $\downarrow$ & SPKSIM (\%) $\uparrow$
              & LSE-C $\uparrow$ & LSE-D $\downarrow$ \\
    \midrule
      GT                & 2.33  & 56.18 & 7.80 & 6.85 \\
      \midrule
      SA+FFN (base)    & 1.52  & 47.12  & 3.17 & 11.25 \\
      \midrule
      +MDR (Single Linguistic Memory)  & 1.43 & 46.53 & 8.17 & 6.59 \\
      +MDR (Single Speaker-Style Memory)  & 1.58 & 51.92 & 8.13 & 6.64 \\
      +MDR (Dual-context Memories)  & 1.39  & 51.49 & 8.18 & 6.56 \\
       RTG + MDR (Dual-context Memories)    & \textbf{1.37} & \textbf{52.05}
                        & \textbf{8.23} & \textbf{6.50} \\
      
      \bottomrule
    \end{tabular}
  \end{table}

\subsection{Ablation on Flow Guidance Strategies}
\label{AblationGuidance}
To evaluate the ability of Rhythm-anchored Trajectory Guidance to preserve audiovisual alignment, we progressively increased the scale and compared RTG with the conventional classifier-free guidance strategy used in the official AlignDiT implementation~\cite{AlignDiT}. 
As shown in Figs.~\ref{fig:joint} (LRS3 results) and~\ref{fig:joint_celeb} (CelebV-Dub results), increasing text guidance generally reduced WER on both benchmarks. However, under conventional guidance, stronger text guidance progressively impaired audiovisual alignment, as indicated by lower AVSync and LSE-C scores and a substantial increase in LSE-D at larger guidance scales. In contrast, RTG preserved or improved AVSync and considerably limited the deterioration in LSE-C and LSE-D while retaining comparable WER improvements. This behavior is consistent with the design of RTG. 
Rather than relying on simple linear extrapolation, RTG computes hierarchical multimodal corrections at a visual-only predictive midpoint and injects them into the fully conditioned velocity field, thereby reducing the tendency of stronger text guidance to override the visual timing established by lip motion. 
Together, these results support the intended protective role of RTG in strengthening linguistic adherence while mitigating audiovisual alignment drift.

\subsection{Ablation on UltraDub Components}
\label{AblationComponents}
Tab.~\ref{tab:lrs3_ab} isolates the contribution of each component on the LRS3 benchmark. 
All architectural variants are trained as separate models using the same training configuration. 
The SA+FFN backbone achieves a low WER but exhibits weak audiovisual alignment. 
Introducing either memory component into MDR substantially improves
synchronization. 
The linguistic memory favors content accuracy, whereas the speaker-style memory improves speaker similarity, revealing the limitations of optimizing either context alone. 
MDR (Dual-context Memories) retrieval better balances these objectives. 
Adding RTG consistently improves all four metrics and yields the best overall results. These gains are obtained without retuning the guidance scales, indicating that RTG provides additional benefits under the same nominal guidance settings. 
Overall, the results validate the contributions of visually guided dual-memory retrieval and rhythm-anchored trajectory guidance.

\begin{table}[th]
  \centering
  \caption{Results on the DiverseDub benchmark. Arrows indicate the preferred direction; the best result among dubbing
  methods is shown in bold. 
  AVSync is the mean cosine similarity between AV-HuBERT features of generated and GT audiovisual pairs; higher is better.}
  \label{tab:diversedub_results_AVSync}
  \resizebox{\linewidth}{!}{%
  \begin{tabular}{lcccccc}
  \toprule
  Method & WER (\%) $\downarrow$ & SPKSIM (\%) $\uparrow$
         & EmoSIM (\%) $\uparrow$ & LSE-C $\uparrow$ & LSE-D $\downarrow$ & AVSync $\uparrow$ \\
  \midrule
  GT               &  1.07 & 69.92 & 100.00 & 6.64 & 7.83 & 1.000 \\
  \midrule
  ProDubber        & 13.86 & 25.17 & 69.29 & 2.40 & 11.87 & 0.175 \\
  VoiceCraftDub   & 84.68 & 41.26 & 72.90  & 4.89 & 9.60 & 0.248 \\
  AlignDiT         & 20.61 & 49.78 & 80.09  & 6.33 & 8.15 & 0.344 \\
  InstructDubber   & \textbf{8.38} & 23.87 & 67.72  & 1.85 & 12.47 & 0.129 \\
  CoSyncDiT   & 12.01 & 56.32 & 81.72  & 6.88 & 7.86 & {0.365} \\

  \midrule
  \textbf{UltraDub (Ours)}
                   & 
                   {9.60} & \textbf{61.15} & \textbf{82.93}
                   & \textbf{7.13} & \textbf{7.56} & \textbf{0.370} \\
  \bottomrule
  \end{tabular}%
  }
  \end{table}

\begin{table}[th]
    \centering
    \caption{Comparison with dubbing methods on the LRS3-Cross benchmark. Best results among dubbing methods are shown in bold. Note that EmoSIM is not evaluated here, as LRS3 predominantly features neutral TED talks with limited emotional variation. AVSync is the mean cosine similarity between AV-HuBERT features of generated and GT audiovisual pairs; higher is better.}
    \label{tab:lrs3_cross_results_AVSync}
    \begin{tabular}{l|ccccc}
      \toprule
      Methods & WER (\%) $\downarrow$ & SPKSIM (\%) $\uparrow$
              & LSE-C $\uparrow$ & LSE-D $\downarrow$ & AVSync $\uparrow$ \\
      \midrule
      GT                & 2.33  & 56.18 & 7.80 & 6.85 & 1.000 \\
      \midrule
      ProDubber         & 3.41  & 24.27 & 4.27 & 10.12 & 0.220 \\
      VoiceCraftDub     & 14.85 & 35.17 & 6.49 & 8.08 & 0.568 \\
    AlignDiT          & 1.40  & 51.46 & 7.91 & 6.78 & 0.751 \\
      InstructDubber    & 2.39  & 24.16 & 3.29 & 11.12 & 0.172 \\
      CoSyncDiT    & 2.09  & 47.53 & 7.83 & 6.88 & 0.731 \\
      \midrule
      \textbf{UltraDub (Ours)}
                        & \textbf{1.37} & \textbf{52.05}
                        & \textbf{8.23} & \textbf{6.50} & \textbf{0.756} \\
      \bottomrule
    \end{tabular}
  \end{table}

\subsection{Additional AVSync Results}
\label{AdditionalAVSync}
To further evaluate audiovisual consistency, we use AVSync~\cite{AlignDiT}, which measures the mean cosine similarity between AV-HuBERT features extracted from generated and ground-truth audiovisual pairs. 
The results are reported in Tabs.~\ref{tab:diversedub_results_AVSync} and~\ref{tab:lrs3_cross_results_AVSync}. 
A higher AVSync score indicates stronger correspondence between the generated speech and visual content. While the previously reported WER, SPKSIM, EmoSIM, LSE-C, and LSE-D results remain unchanged, UltraDub achieves the highest AVSync score among all dubbing methods on both DiverseDub and LRS3, with scores of 0.370 and 0.756, respectively. It outperforms the best-performing baseline on both benchmarks, demonstrating more accurate and consistent audiovisual alignment.

\subsection{Subjective MOS Evaluation} 
\label{SubjectiveMOS}
\begin{figure}[tb]
  \centering
  \includegraphics[width=1.0\linewidth]{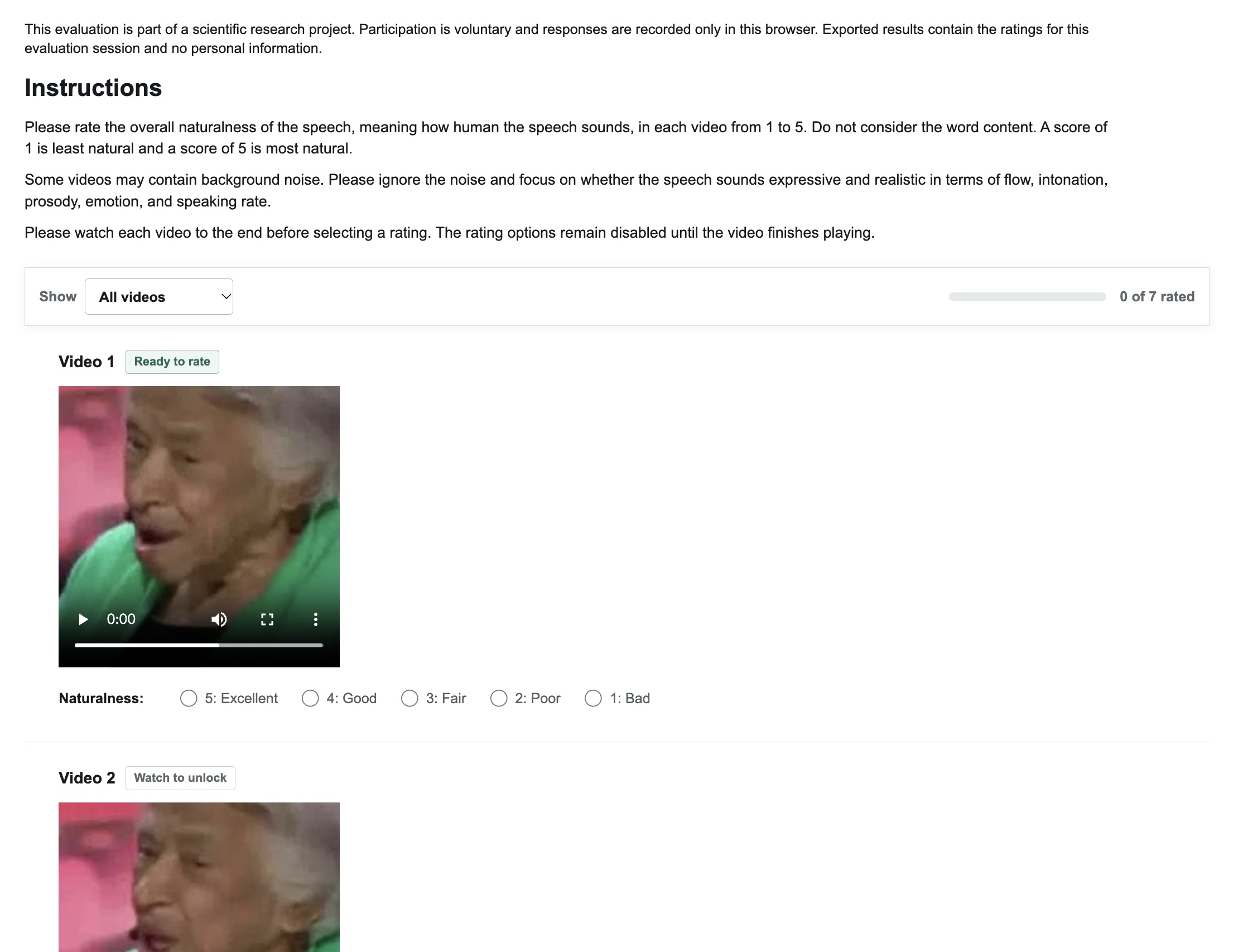}
  \caption{Instructions and an example item for the human listening test
of overall speech naturalness.}
  \label{fig:MOS_Eva}
\end{figure}

Following prior work, we conduct a five-point mean opinion score evaluation involving 20 listeners and 30 randomly selected samples from LRS3-Cross. 
The instructions and an example item used in the listening test are shown in Fig.~\ref{fig:MOS_Eva}. 
MOS-N evaluates speech naturalness, whereas MOS-S measures perceptual similarity to the reference speaker. As shown in Tab.~\ref{tab:subjective_lrs3}, the subjective results on LRS3-Cross further confirm the perceptual advantages of UltraDub. UltraDub achieves the highest naturalness score among all dubbing methods, obtaining 3.96 $\pm$ 0.13, compared with 3.86 $\pm$ 0.12 for CoSyncDiT and 3.82 $\pm$ 0.14 for AlignDiT. For speaker similarity, UltraDub achieves 4.01 $\pm$ 0.12, surpassing all competing methods. These results indicate that UltraDub generates more natural speech while more faithfully preserving the identity of the reference speaker.

\subsection{Limitations and Social Impact}
\label{Limitations}

\begin{table}[!t]
  \centering
  \caption{Subjective evaluation results on LRS3-Cross for naturalness (MOS-N) and speaker similarity (MOS-S). }
  \label{tab:subjective_lrs3}
    \begin{tabular}{l cc}
      \toprule
      \textbf{Method}
      & \textbf{Naturalness (MOS-N)} $\uparrow$
      & \textbf{Similarity (MOS-S)} $\uparrow$ \\
      \midrule
      Ground Truth
      & 4.15 $\pm$ 0.12
      & 4.01 $\pm$ 0.13 \\
      \midrule
      ProDubber
      & 3.10 $\pm$ 0.17
      & 2.55 $\pm$ 0.13 \\
      VoiceCraftDub
      & 3.31 $\pm$ 0.09
      & 3.15 $\pm$ 0.15 \\
      InstructDubber
      & 3.36 $\pm$ 0.12
      & 2.60 $\pm$ 0.14 \\
      AlignDiT
      & 3.82 $\pm$ 0.14
      & 3.99 $\pm$ 0.13 \\
      CoSyncDiT
      & 3.86 $\pm$ 0.12
      & 3.98 $\pm$ 0.15 \\
      \midrule
      \textbf{Ours}
      & \textbf{3.96 $\pm$ 0.13}
      & \textbf{4.01 $\pm$ 0.12} \\
      \bottomrule
    \end{tabular}
\end{table}

By generating speech that preserves speaker identity and remains synchronized with silent video, UltraDub can facilitate film post-production, accessibility services, and speech restoration for videos with missing or corrupted audio. 
However, this capability also introduces risks of unauthorized voice cloning, impersonation, and deceptive audiovisual content. In particular, malicious users could combine manipulated scripts with a target speaker's reference speech and facial motion to generate statements that the depicted person never made. The present work focuses on generation quality and does not provide a complete solution for detecting or preventing such misuse. Responsible deployment should therefore incorporate informed consent procedures, explicit disclosure of synthesized content, provenance tracking, access controls, and human oversight. Developing reliable watermarking and misuse-detection mechanisms remains an important direction for future work.

\end{document}